\documentclass[journal]{IEEEtai}

\usepackage[colorlinks,urlcolor=blue,linkcolor=blue,citecolor=blue]{hyperref}

\usepackage{color,array}

\usepackage{graphicx}
\usepackage{amsmath}

\begin{document}

\title{DEM-NeRF: A Neuro-Symbolic Method for Scientific Discovery through Physics-Informed Simulation}

\author{Wenkai Tan, Alvaro Velasquez, and Houbing Song, \IEEEmembership{Fellow, IEEE}
\thanks{This material is based upon work supported by the National Science Foundation under Grant No. 2229155. The opinions, findings, and conclusions, or recommendations expressed are those of the author(s) and do not necessarily reflect the views of the National Science Foundation.}
\thanks{Wenkai Tan and Houbing Song are with the Department of Information System, University of Maryland, Baltimore County, Baltimore, MD 21250 USA (e-mail: wtan1@umbc.edu; hsong@umbc.edu).}
\thanks{Alvaro Velasquez was with University of Colorado, Boulder, CO 80309 USA. He is now graduated and received his Ph.D degress.}
\thanks{This paragraph will include the Associate Editor who handled your paper.}}

\markboth{IEEE Transactions on Artificial Intelligence, Vol. 00, No. 0, Month 2020}
{Tan \MakeLowercase{\textit{et al.}}: DEM-NeRF}

\maketitle

\begin{abstract}
Neural networks have emerged as a powerful tool for modeling physical systems, offering the ability to learn complex representations from limited data while integrating foundational scientific knowledge. In particular, neuro-symbolic approaches that combine data-driven learning, the “neuro”, with symbolic equations and rules, the “symbolic”, address the tension between methods that are purely empirical, which risk straying from established physical principles, and traditional numerical solvers that demand complete geometric knowledge and can be prohibitively expensive for high-fidelity simulations.
In this work, we present a novel neuro-symbolic framework for reconstructing and simulating elastic objects directly from sparse multi-view image sequences, without requiring explicit geometric information. Specifically, we integrate a neural radiance field (NeRF) for object reconstruction with physics-informed neural networks (PINN) that incorporate the governing partial differential equations of elasticity. In doing so, our method learns a spatiotemporal representation of deforming objects that leverages both image supervision and symbolic physical constraints. To handle complex boundary and initial conditions, which are traditionally confronted using finite element methods, boundary element methods, or sensor-based measurements, we employ an energy-constrained Physics-Informed Neural Network architecture. This design enhances both simulation accuracy and the explainability of results.
Our experimental results showcase high-quality reconstructions and reliable deformation predictions in hyperelastodynamics and loading scenarios. These findings highlight the transformative potential of neuro-symbolic approaches in computational mechanics and suggest broader applications in real-time physics simulation, materials design, biomedical technology, and sustainable energy solutions. By unifying empirical observation with theoretical abstraction in a single trainable system, our work marks a significant advancement in using machine learning for scientific discovery.
\end{abstract}

\begin{IEEEImpStatement}
Simulating hyperelastic deformation in real-world solids remains a challenging task due to unknown boundary conditions, high computational costs, and incomplete geometric information. This work proposes DEM-NeRF, a neuro-symbolic framework that integrates NeRF with physics-informed deep energy methods to reconstruct and simulate solid deformation directly from image sequences. Unlike traditional numerical solvers or purely data-driven networks, our approach unifies empirical visual cues with physical constraints, enabling real-time, accurate predictions of deformation without labeled data or specialized hardware. DEM-NeRF represents a significant step toward interpretable, physics-grounded machine learning for computational mechanics, with broad implications for biomedical imaging, soft robotics, and industrial digital twins.
\end{IEEEImpStatement}


\section{Introduction}
\label{sect:introduction}

\IEEEPARstart{S}{cientific} progress has traditionally been driven by two foundational engines: empirical observation and theoretical abstraction \cite{dvzeroski2007computational}. Their interplay has yielded increasingly sophisticated models of physical phenomena, culminating in computational simulations that span scales from subatomic interactions to galactic processes \cite{lu2019simulations, taylor2016time}. More recently, the integration of advanced machine learning methods with well-established scientific principles, the neuro-symbolic approach, has begun to reshape how these simulations are conceptualized and executed \cite{karniadakis2021physics, raissi2019physics}. In this paradigm, the “neuro” component exploits the pattern recognition and representational capabilities of neural networks, while the “symbolic” component emphasizes domain-specific equations and symbolic rules, which have long provided the theoretical cornerstone of scientific discovery \cite{votsis2023neuro}.


By fusing data-driven learning with strict adherence to physical laws, neuro-symbolic techniques overcome limitations found in both purely empirical models and traditional numerical methods. Empirical models often violate known physical constraints, while numerical solvers can be computationally intensive and difficult to apply to highly nonlinear, multi-scale phenomena \cite{quarteroni2008numerical, larson2015finite}. This synergy has proven essential in fields such as climate modeling, astrophysics, fluid mechanics, and materials science, where increasing complexity challenges even the most powerful high-performance computing infrastructures \cite{towns2014xsede, post2006opportunities}. By embedding equations such as conservation laws and constitutive models directly into neural network architectures and loss functions, neuro-symbolic approaches maintain consistency with physical principles while reducing dependence on large labeled datasets \cite{karniadakis2021physics, raissi2019physics}. These frameworks improve interpretability and reliability, offering strong potential for breakthroughs in real-time simulation, materials discovery, and other scientific domains \cite{lu2024surveying, hitzler2022neuro}. This convergence of machine learning and physics represents a pivotal advancement in computational science, expanding our ability to address complex and previously intractable problems.

Neuro-symbolic frameworks, positioned at the intersection of the data-driven adaptability of machine learning and the foundational rigor of physical laws, have demonstrated remarkable potential in addressing the complexities of hyperelastodynamics. Traditional computational approaches such as the finite element method (FEM) \cite{taylor2013finite}, boundary element method (BEM) \cite{brebbia1994boundary}, and material point method (MPM) \cite{hu2019taichi} are highly accurate but often rely on extensive expert knowledge, complete geometric information, and comprehensive boundary conditions. In contrast, purely data-driven methods can produce visually convincing results but risk deviating from established scientific principles due to a lack of explicit physical constraints. Neuro-symbolic techniques seek to unify these two methodologies by embedding fundamental conservation laws, energy equations, and boundary conditions directly into neural architectures. This integration ensures physically plausible solutions and supports real-time performance, which is crucial for modern engineering and scientific applications. Building on neural radiance fields (NeRF) and physics-informed neural networks (PINNs), these hybrid approaches offer a promising new direction for scientific discovery by enabling precise, scalable, and interpretable simulation of hyperelastic phenomena across diverse real-world scenarios.

Based on the recently proposed view synthesis NeRF method \cite{mildenhall2021nerf}, several methods have been proposed on shifting from static scene modeling to capturing dynamic phenomena, allowing for more realistic representations of time-varying objects and environments. Dynamic NeRF \cite{pumarola2021d} extends the original framework by introducing temporal dependencies, enabling the reconstruction of objects undergoing motion or deformation. Building on these advancements, researchers have proposed Physics-Informed Neural Radiance Fields (PI-NeRF) \cite{chu2022physics}, which incorporate governing physical equations directly into the NeRF training process. A key application of PI-NeRF lies in fluid dynamic reconstruction. PI-NeRF leverages physical priors to constrain the radiance and density fields, ensuring consistency with fluid flow equations and reducing the reliance on large amounts of training data.
By embedding domain-specific constraints, such as velocity, PI-NeRF goes beyond purely data-driven approaches and enforces physical realism in reconstructed scenes. However, the velocity field learned in fluid dynamic reconstruction is a weak-boundary or boundary-free field. The boundary condition in this method limits the PI-NeRF's generalizability and transferability on learning other physics phenomena.

Inspired from the PI-NeRF, Li et al. proposed PAC-NeRF \cite{li2023pac} to solve physics phenomena with hard-boundary conditions, such as solid deformation. A computational physics method, MPM, is used in PAC-NeRF to solve real-time solid deformation. Although MPM is more efficient than FEM and pure particle methods, large-scale and real-time simulations can become computationally expensive. We aim to overcome these difficulties, obtain 3D objects without manual labels, handle solid scenes with unknown hard-boundary conditions, and reduce the computational cost in real-time simulations. We propose the first neuro-symbolic method, DEM-NeRF, for real-time solid deformation reconstruction and prediction.

\section{Background and related work}

\label{sect:background}
Ongoing efforts aim to enhance the capability of using neural networks to solve real-world tasks with domain prior knowledge efficiently. In line with this, recent studies explore strategies for combining computer vision methods with computational physics knowledge to simulate real-world physical phenomena. Since these studies directly relate to our own investigation, this section briefly introduces three main topics: Solid's Deformation Reconstruction Methods, PINN and its related applications, and Neural Representation of Solids.

\subsection{Solid's Deformation Reconstruction Methods}
Several established approaches exist for reconstructing solid deformations, ranging from digital image correlation (DIC) \cite{sutton2009image} and holographic interferometry \cite{schumann2013holographic} to active sensing techniques involving specialized hardware and lighting setups. For instance, DIC tracks surface speckle patterns across multiple images to measure deformations under load. Its non-intrusive nature makes DIC advantageous for capturing full-field strain distributions in both laboratory and industrial settings. In this work, we focus on solid's deformation reconstruction using RGB images to reduce the reliance on specialized hardware and experimental setups. Early work \cite{sutton2009image} uses non-contact optical methods to measure surface deformation, displacement, and strain in materials under load. It works by comparing images of a sample taken before and after deformation. By correlating the shift and distortion of speckle patterns present on the sample’s surface between images, DIC algorithms compute full-field displacement maps. Extending this direction, the RGB-D approach \cite{franco2019rgb} uses a low-cost Kinect V2 camera, which provides both color and depth data for three-dimensional displacement measurements. By applying digital image correlation (DIC) principles to these depth images, the method can measure deformations and displacements comparably to higher-end, specialized systems.
More recent methods leverage standard RGB images and deep learning pipelines to infer 3D displacements without relying on complex experimental configurations \cite{shao20243d}, thus broadening the applicability of deformation reconstruction in real-world scenarios.

\subsection{Physics Informed Neural Network}
PINNs embed partial differential equations (PDEs) in the neural network training process. They incorporate PDE residuals and boundary conditions as part of the loss function. Early PDE solvers used multilayer perceptrons (MLPs) for direct approximation of PDE solutions. These approaches harnessed universal approximation properties. Lagaris et al. \cite{lagaris1998artificial} introduced MLP-based solvers under controlled settings. Raissi et al. \cite{raissi2019physics} formalized PINNs and enforced PDE residuals and boundary conditions to address forward and inverse problems with minimal labeled data. Recent work has explored multi-physics simulations and high-dimensional PDEs under the PINN framework. Karniadakis et al. \cite{karniadakis2021physics} reviewed recent advanced techniques for embedding physics into machine learning, and showcased a range of physics-informed learning applications for forward and inverse problems. PINNs found success in modeling fluid dynamics. Raissi et al. \cite{raissi2019deep} investigated vortex-induced vibrations with reduced data requirements. PINNs also addressed thermal dynamics and heat transfer in Navier-Stokes flow regimes. Jin et al. \cite{jin2021nsfnets} solved thermal processes with sparse or noisy measurements. These applications demonstrated the ability of PINNs to merge domain knowledge and data-driven methods. PINNs improved generalization and interpretability compared to purely data-driven models. They offered a unified framework that bridged PDE solvers and neural network methodologies. They also enhanced solution quality and reduced computational expense.

\subsection{Neural Representation of Solid Geometry}
NeRF \cite{mildenhall2021nerf} represents 3D scenes as continuous functions. They map spatial coordinates and viewing directions to color and density. The original NeRF work focused on static scenes and reconstructed them from sparse camera views. NeRF achieved photorealistic results by using volume rendering and positional encoding. Building upon static NeRF, dynamic NeRF approaches generalize the concept to model temporal variation. They incorporate an additional dimension for time \cite{pumarola2021d}. They model time-varying geometry and appearance. These methods learn a motion field or deformation to handle scene dynamics. This representation allows rendering from arbitrary viewpoints at any moment in time. It also captures non-rigid motions, such as waving cloth or moving fluids \cite{chu2022physics}. NeRF-based methods have been applied to fluid reconstruction. Fluid scenes involve complex density changes and transient motion. A dynamic neural representation can learn to map spatiotemporal coordinates to fluid density and color. This approach bypasses traditional physics-based solvers. Neural representations can capture realistic smoke or liquid behaviors directly from image sequences. NeRF-based methods can also assist in solid deformation reconstruction \cite{li2023pac, feng2024pie}. These methods learn to represent changing surface geometry and texture, and use material point method (MPM) as a computational physics tool to solve solid’s deformation. Researchers combine NeRF reconstructions with physics priors to improve the reconstruction quality and the simulation accuracy. They achieve accurate results even under challenging real-world conditions \cite{li2023pac}. They reduce reliance on specialized hardware setups.

\section{Methodology}

\label{sect:methodology}

\subsection{Problem Definition}
\begin{figure}[h]
  \centering
  \includegraphics[width=1\linewidth]{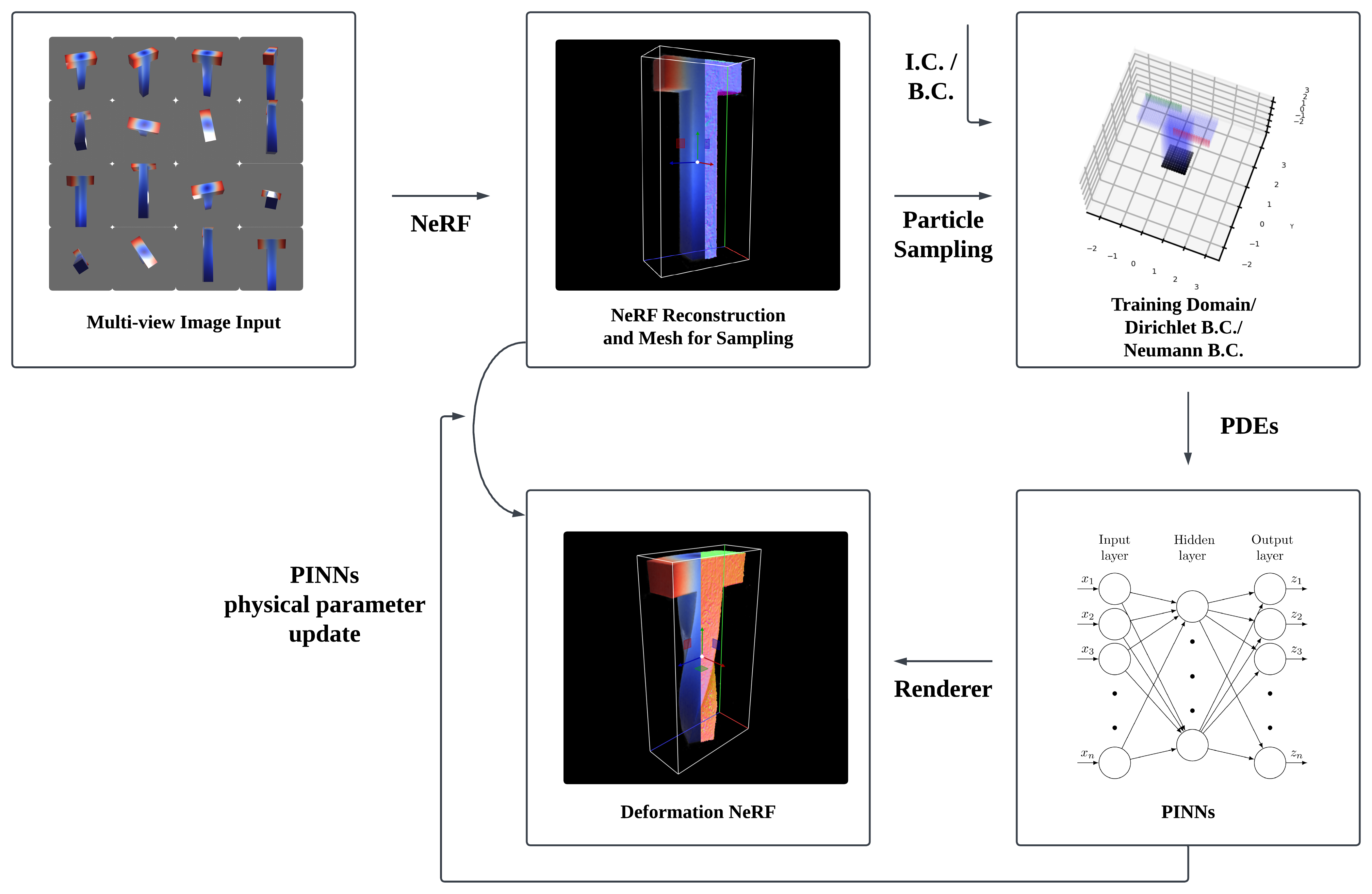}
  \caption{The pipeline of DEM-NeRF.}
  \label{fig:DEM-NeRF_pipeline}
\end{figure}

To solve the real-time solid reconstruction and current limitations, we propose DEM-NeRF.
Figure~\ref{fig:DEM-NeRF_pipeline} illustrates the pipeline of our proposed DEM-NeRF method. In DEM-NeRF, we will first perform a 3D object reconstruction using NGP-NeRF \cite{muller2022instant}. The positional and texture information will be encoded using NGP. The NGP method also provides NeRF to Mesh conversion. Second, we use sampling techniques to obtain a particle cloud representation of the 3D hyperelastic Neo-Hookean solid body $\mathcal{B}$. This method can adopt any sampling techniques, including random sampling, Poisson disc sampling, and uniformed mesh sampling. It is noted that while Poisson disc sampling outperforms other methods on complex boundary 3D shapes, uniform mesh sampling performs well on simple 3D shapes such as cuboids. After we obtain the particle-could representation of the 3D object, the particle set will pass into DEM for real-time interactive hyperelastic deformation predictions.

\subsection{Neural Scene Representation for Solid}
\label{sect:methodNeRF}
Building upon the universal approximation theorem, our approach utilizes neural networks to characterize the spatial properties of a deformable object. Concretely, we introduce two neural networks, $F_{\mathrm{NeRF}}$ and $F_{\mathrm{DEM}}$ in Eq.~\ref{eq_F}, which approximate the continuous radiance field and the solid’s displacement field, respectively.

\begin{equation}
    \label{eq_F}
    \begin{split}
        F_{\mathrm{NeRF}} &: (x,y,z) \rightarrow (\mathbf{c},\sigma), \\
        F_{\mathrm{DEM}} &: (x,y,z) \rightarrow (u).
    \end{split}
\end{equation}

Here, $(x,y,z)$ encodes the three-dimensional structure of the object, while the output of $F_{\mathrm{NeRF}}$ consists of the emitted color $\mathbf{c}$ and the volume density $\sigma$. In parallel, the output of $F_{\mathrm{DEM}}$ produces the displacement vector $u$.

By using the radiance field, we gain valuable insight into lighting conditions, object geometry, and how sampling particles are distributed in space. Meanwhile, the displacement field captures hidden elastic energy-essential for calculating strain and internal or external forces within hyperelastic frameworks. Although $F_{\mathrm{NeRF}}$
can be trained via differentiable volumetric rendering from multi-view image sequences, $F_{\mathrm{DEM}}$ depends on physics-based constraints, utilizing the particle distribution in the training domain for indirect supervision. Notably, the relevant physical laws involve strain energy $\Psi$, which can be seen as a functional transformation of the volume density $\sigma$ under the neo-Hookean model’s strain energy formulation.

Informed by the traditional 3D rendering techniques, Mildenhall et al. introduced the neural radiance field to express particles in space through a combination of volume density $\sigma$ and directionally emitted radiance \cite{mildenhall2021nerf}. In this framework, $\sigma(x)$ can be interpreted at spatial location $x$ with the camera ray $\textbf{r}(s)$. The expected color can be calculated as:

\begin{equation}
    \label{eq_renderer}
    \begin{split}
        & C(\textbf{r}) =  \int^{s_f}_{sn}T(s)\sigma(\textbf{r}(s))\textbf{c}(\textbf{r}(s),\omega)ds,\\
        & T(s)=\text{exp}\left(-\int^s_{s_{n}}\sigma(\textbf{r}(s))ds\right)
    \end{split}
\end{equation}.

This volume rendering methodology enables us to directly estimate scene geometry, appearance, and particle density from a continuous radiance field, which in turn serves as the foundation for our object reconstruction and mesh extraction strategies.

\subsection{Particle Sampling}
\label{sect:methodParticle}
The proposed DEM-NeRF method can adopt any particle sampling method. In this work, we discuss three techniques.

The random sampling technique involves placing a set of randomly generated particles into a 3D object in rectangular coordinates. We use the representation Eq.(~\ref{eq_randomSample}) to describe its process. This process is similar to Monte Carlo simulation.
\begin{equation}
    \label{eq_randomSample}
    \{x,y,z\}=\{x,y,z\}_{\min}+U(0,1)\cdot (\{x,y,z\}_{\max}-\{x,y,z\}_{\min})
\end{equation}

Poisson disc sampling has a tightly-packed particle distribution. Every two particles will have a random distance between [r, 2r]. The particles are uniformly distributed in a spherical shell instead of a rectangular cube.

Uniform mesh sampling divides a 3D object into small cubes in rectangular coordinates, then places particles to fill all cubes. Each subsequent particle along an axis in rectangular coordinates is placed at a fixed distance.
\begin{equation}
    \label{eq_uniformedMeshSample}
    x^{(k)}_{i_{k}}=a_{k}+\frac{(i_k-1)(b_{k}-a_{k})}{N_{k}-1}
\end{equation}

Each of these sampling techniques has its own advantages and disadvantages. We will discuss the comparison in \ref{sect:evalParticle}.

\subsection{Deep Energy Method}
\label{sect:methodDEM}
In this work, we focus on a Neo-Hookean solid. A homogeneous, isotropic, and nonlinear hyperelastic Neo-Hookean solid body $B$ is described by a set of particles or sampling points which is sampled by Poisson disc sampling or uniformed mesh sampling. This body is bounded by an elastic surface $\partial B$ in its initial condition.  A time-dependent motion between deformation states can be described by a mapping function Eq.(~\ref{eq_mapping}) that maps particles from the initial configuration to the current one.
\begin{equation}
    \label{eq_mapping}
    \phi\left(X,t\right)=u+X
\end{equation}
where $u$ denotes the displacement vector. A deformation gradient that describes the gradient of the particles' motion with respect to the initial condition can be calculated as
\begin{equation}
    \label{func_defGrad}
    F=\mathrm{Grad} \phi(X).
\end{equation}

For this compressible Neo-Hookean solid, its strain energy is represented by the following
\begin{equation}
    \label{eq_strainEnergy}
    \Psi(I_1,J) = \frac{1}{2}\mu (I_1-3)-\mu ln(J)+\frac{1}{2}\lambda(ln(J))^2
\end{equation}
where the first invariant $I_1=tr\left(F^\top F\right)$, the second invariant $J=det\left(F\right)$. $\mu,\lambda$ are Lamé parameters that describe the characteristics of the material; they can be calculated as follows
\begin{equation}
    \label{eq_Lame}
    \mu = \frac{E}{2(1+\nu)}, \lambda = \frac{E\nu}{(1+\nu)(1-2\nu)}
\end{equation}
with the Young’s modulus $E$ and the Poisson’s ratio $\nu$.

We can define the total potential energy $\Pi$ as the sum of the stored strain energy and the potential energy of the external forces, which denote as follows
\begin{equation}
    \label{eq_potentialEnergy}
    \Pi(\phi)=\int_{B}\Psi dV-\int_{B}f_b\cdot \phi dV-\int_{\Gamma_t}\overline{t}\cdot \phi dA
\end{equation}
where $f_b$ is the body force vector, and $\bar{t}$ is the traction boundary condition.
A quasi-static equilibrium can describe the state of a solid without motion. For any deformation states after motion, the energy function described in Eq.(~\ref{eq_potentialEnergy}) needs to be minimized to fulfill the static equilibrium.

\begin{figure*}[ht]
    \centering
    \includegraphics[width=1\linewidth]{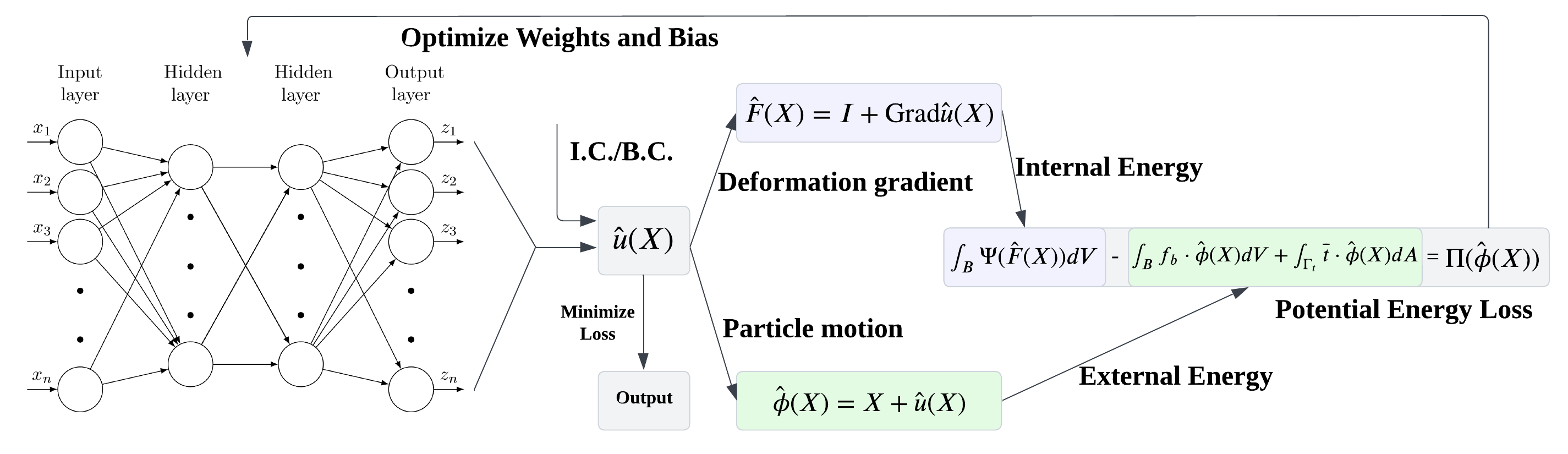}
    \caption{Deep Energy Method overview.}
    \label{fig:DEM_net}
\end{figure*}

In Deep Energy Method (DEM) \cite{nguyen2020deep}, the neural network solves quasi-static equilibrium problems by acting as a global shape function that represents the displacement throughout the body. The neural network architecture is shown in Figure~\ref{fig:DEM_net}. The network encodes particle positions into displacement vectors. This representation models the solid’s deformation. Rao et al. discussed that the neural network can also be trained to fit the boundary condition \cite{rao2021physics}. We let the displacement vector be equal to the output of the network, which is denoted as
\begin{equation}
    \label{eq_DEMout_displace}
    \hat{u}(X) = z(X).
\end{equation}

With the intermediate displacement vector $\hat{u}$, we can calculate the deformation gradient as
\begin{equation}
    \label{eq_deform_grad}
    \hat{F}(X) = I+\mathrm{Grad}\hat{u}(X).
\end{equation}
Based on the DEM's architecture, a particle's displaced position is needed to calculate the potential energy. The particle's displaced position is obtained by adding the displacement vector to the original position:
\begin{equation}
    \label{eq_motion}
    \hat{\phi}(X) = X+\hat{u}(X).
\end{equation}

As shown in Figure~\ref{fig:DEM_net}, the total loss is formed with potential energy loss and boundary loss. For all sampled particles $X$ present in the Neo-Hookean solid body $\mathcal{B}$, we use $\{X^i_\Pi\}^{N_\Pi}_{i=1}$ to represent the set of domain particles, and $\{X^i_u\}^{N_u}_{i=1}$ to represent the set of boundary particles. The total loss of the DEM can be calculated as
\begin{equation}
    \label{eq_DEMtotalLoss}
    \begin{split}
        \Theta^* = & \textbf{argmin}_\Theta  ( \Pi(\hat{\phi}(\{X^i_\Pi\}^{N_\Pi}_{i=1},\Theta)) \\
        & +W_u\mathrm{MSE}_u(\{X^i_u\}^{N_u}_{i=1},\Theta) )
    \end{split}
\end{equation}
where $\Theta=\{W^k,b^k\}$ are the trainable parameters and $W_u$ is the PINN weight for boundary loss.
The potential energy loss term is derived from Eq.(~\ref{eq_potentialEnergy}) which denotes as

\begin{equation}
\label{eq_DEMpotentialLoss}
    \begin{split}
        \Pi(\hat{\phi}(\{X^i_\Pi\}^{N_\Pi}_{i=1},\Theta)) &=   \int_{B}\Psi(\hat{F}(\{X^i_\Pi\}^{N_\Pi}_{i=1},\Theta)) dV \\
        &-\int_{B}f_b\cdot \hat{\phi}(\{X^i_\Pi\}^{N_\Pi}_{i=1},\Theta) dV\\
        &-\int_{\Gamma_t}\overline{t}\cdot \hat{\phi}(\{X^i_\Pi\}^{N_Pi}_{i=1},\Theta) dA,
    \end{split}
\end{equation}

\begin{equation}
    \label{eq_DEMboundaryLoss}
    \mathrm{MSE}_u(\{X^i_u\}^{N_u}_{i=1},\Theta)=\frac{1}{N_u}\sum\limits^{Nu}_{i=1}|\hat{u}(X^i_u,\Theta)-u(X^i_u)|^2.
\end{equation}

\section{Evaluation and Discussion}
\label{sect:evaluation}

\subsection{Experiment}
\label{sect:evalSetup}
In this experiment, we mainly utilized the T bar dataset as a showcase. At the default setup, the NeRF can capture not only the texture and location of the targeted object but also the background. Therefore, there are two ways we can extract clear structure from the NeRF. First, we can pass the data through the CLIP method or perform image segmentation to isolate the targeted object’s pixels from the background. Second, we can place the object in the environment with a single color background, then adjust the bounding box during the NeRF training.

The multi-view image sequence with background removal will then pass through the NGP-NeRF to perform a 3D object reconstruction. The reconstructed 3D object will be stored in the NeRF-CUDA renderer. A mesh transformation and uniform particle sampling will be performed within the NeRF renderer to obtain the particle view of the 3D structure.

\begin{figure} [h]
    \centering
    \includegraphics[width=0.6\linewidth]{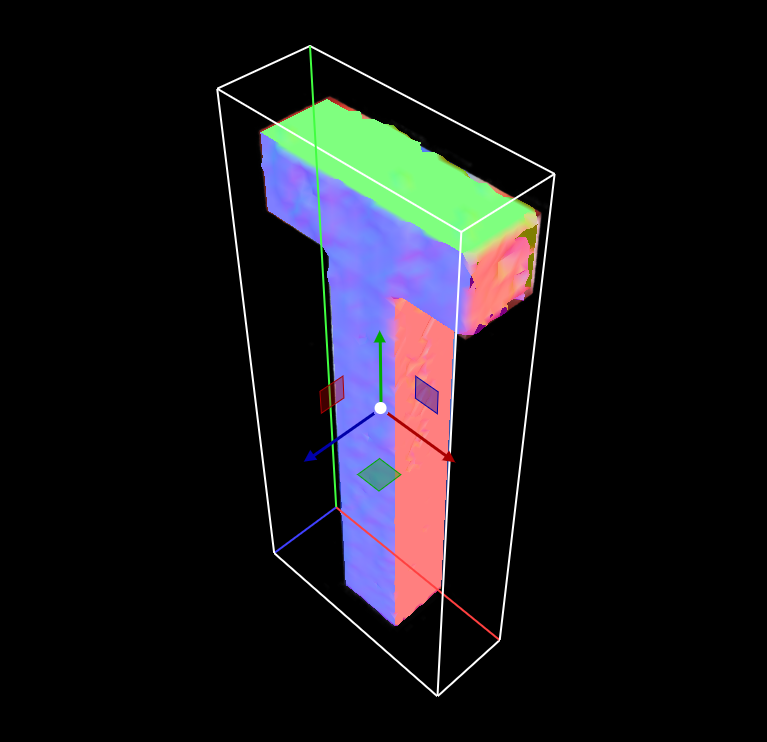}
    \caption{Object reconstruction from undeformed multi-view dataset.}
    \label{fig:undef_TbarMesh}
\end{figure}

\begin{figure} [h]
    \centering
    \includegraphics[width=0.7\linewidth]{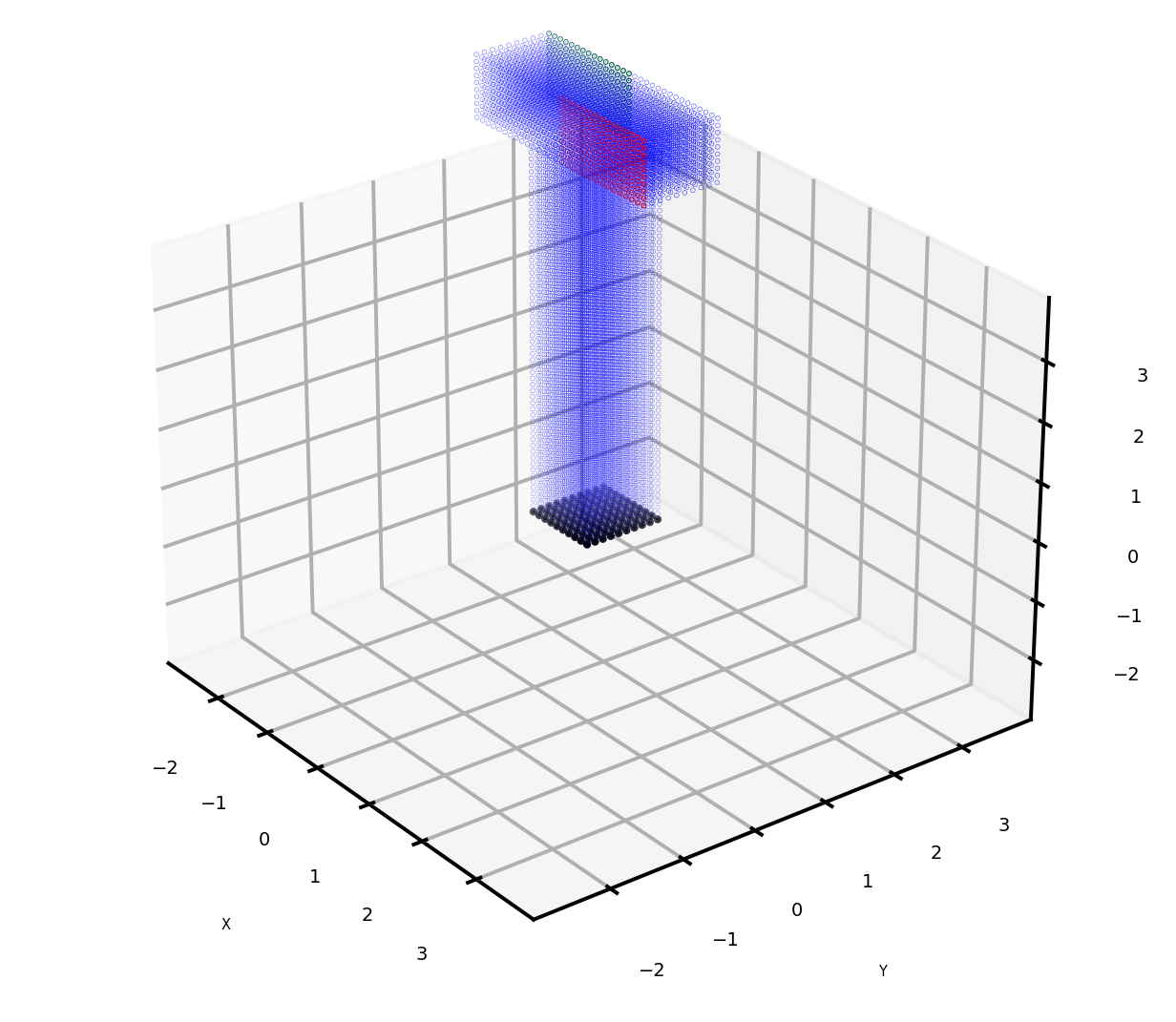}
    \caption{T bar with training domain (blue), Dirichlet Boundary Condition (black), and Neumann Boundary Condition (red, green) setup by using mesh sampling.}
    \label{fig:TbarPart}
\end{figure}


As the visualization shown in Figure~\ref{fig:TbarPart}, a torque formed by two opposing forces (on the front-right and back-left) is applied to the horizontal bar. The bottom of the vertical leg has been fixed. The corresponding regions will form the domain, traction boundary, and displacement boundary. All conditions above lead to the deformation of this hyperelastic solid. 

\begin{figure*}[h]
    \begin{minipage}{0.418\textwidth}
        \centering
        \includegraphics[width=1.2\linewidth]{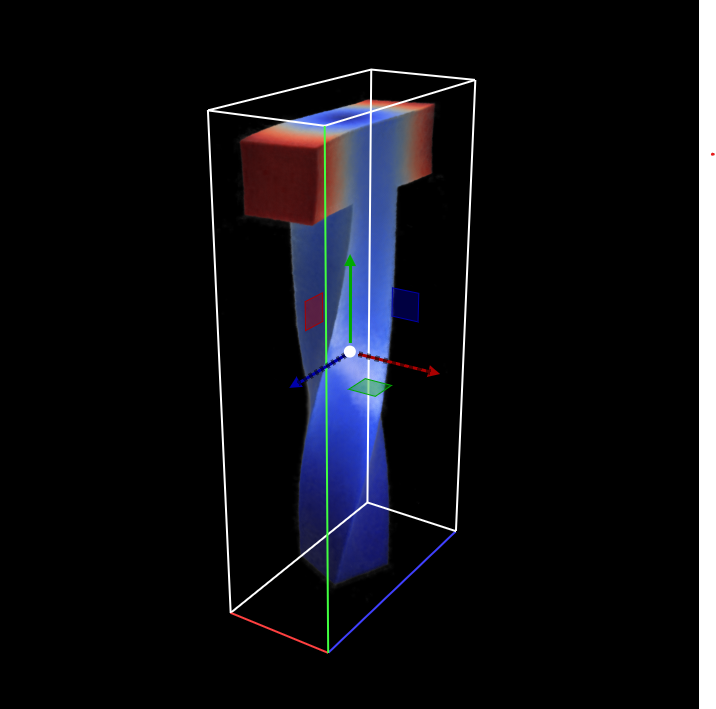}
        \caption{NeRF rendering from DEM output.}\label{fig:def_TbarCol}
    \end{minipage}\hfill
    \begin{minipage}{0.49\textwidth}
        \centering
        \includegraphics[width=1\linewidth]{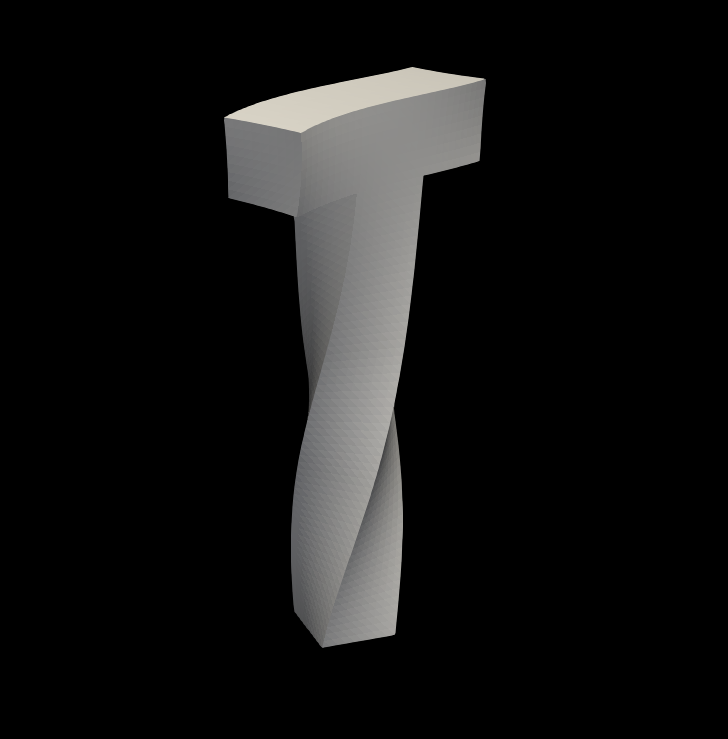}
        \caption{Deformed T bar Ground Truth.}\label{fig:TbarGT}
    \end{minipage}
\end{figure*}

The DEM will be trained with all sampling particles and I.C./B.C. for deformation simulation. In Figure~\ref{fig:def_TbarCol}, we use the same rendering pipeline to obtain the geometry and new views on the deformed structure.

We implemented the DEM-NeRF pipeline using Python and pre-built NGP-NeRF. The neural network for simulation was based on Torch-CPU. Our hardware platform is an AMD Ryzen 7 9800X3D CPU and an NVIDIA RTX4090 GPU. In addition, we performed the same task with different simulation methods. In Table~\ref{table:time_comp}, we demonstrate the significant advantage of using the PINNs-based method in the simulation.

\begin{table*} [t]
\centering
\begin{tabular}{ |p{4cm}||p{4cm}|p{4cm}|  }
 \hline
 \multicolumn{3}{|c|}{Performance Comparison} \\
 \hline
 Methods & Computational time (s) & Rendering time (s) \\
 \hline
 {Full-order FEM \cite{guo2018reduced}} & \centering{1,140} & Not Support \\
 \hline
 PAC-NeRF (MPM) \cite{li2023pac, ceccato2022anura} & Training: $\sim 600$ & NeRF Rendering: $\sim 1$ \\
 \hline
 DEM-NeRF (Ours) & NeRF Training: $\sim 45$ \newline DEM Training: $\sim20$ \newline DEM Predict: $\sim1$ & NeRF Rendering: $\sim 1$\\
 \hline
\end{tabular}
\label{table:time_comp}
\end{table*}

\subsection{Advantage and Disadvantage on Particle Sampling Methods}
\label{sect:evalParticle}
\begin{figure}[h]
    \centering
    \includegraphics[width=0.9\linewidth]{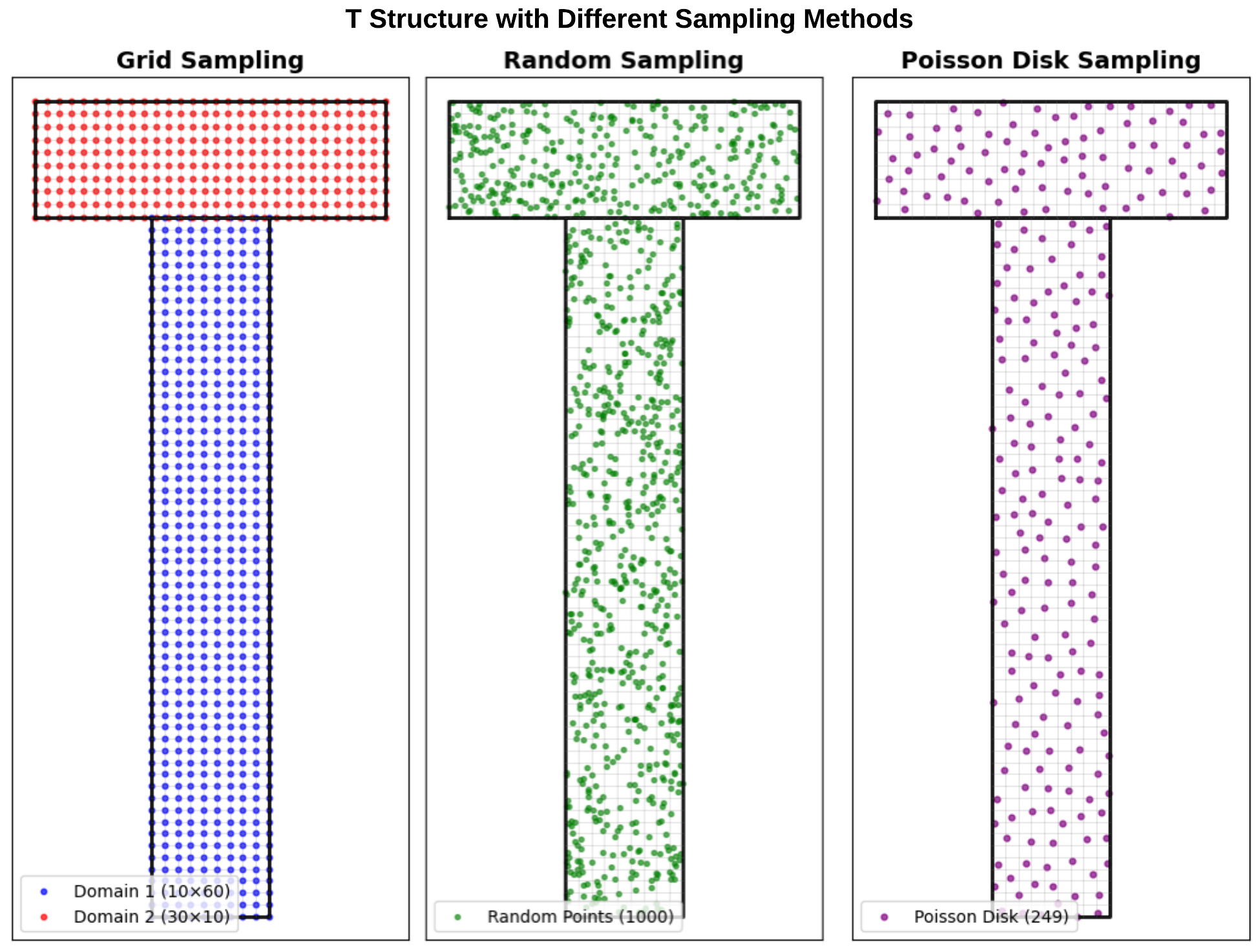}
    \caption{Particle sampling methods on Mesh.}
    \label{fig:particleMethods}
\end{figure}

We have presented three types of particle sampling methods in Section \ref{sect:methodParticle}. In Figure~\ref{fig:particleMethods}, we apply different sampling methods to sample particles within the boundary of a teapot-shaped object.

In the first figure, random sampling generates particle clusters in some regions. These particle clusters can result in irregular distributions of body forces and boundary condition violations. Thus, we recommend using Poisson disc sampling to obtain a naturally distributed particle set. As shown in the second figure, in the case of a 3D structure with complex boundary conditions, Poisson disc sampling allows more particles to be placed on the boundary. This characteristic helps to fulfill the boundary condition requirement for DEM. Thus, Poisson disc sampling can outperform other sampling methods in 3D structures with complex boundary conditions. 
As we discussed in Section~\ref{sect:methodParticle}, a 3D structure such as a cuboid or T-bar, with boundaries (e.g. edge and surface) that align with mesh axes, is naturally suited to uniform mesh sampling. The sampling particles can be perfectly allocated onto the hard-boundary of the structure.

\subsection{Discussion on DEM and PINN}
\label{sect:evalDEMvsPINN}
In the original form of PINN, a first Piola-Kirchhoff stress tensor is inferred from the displacement field in the training domain to fulfill the quasi-static condition which contains a strong form of equilibrium and boundary conditions.
As discussed in Section~\ref{sect:methodDEM}, the computation of the displacement vector and the deformation gradient enables derivation of the first Piola-Kirchhoff stress tensor from the deformation gradient as:
\begin{equation}
    \label{eq_firstPiolaStress}
    P(\hat{F}) = \frac{\partial \Psi}{\partial F}.
\end{equation}
The equilibrium equation is defined as
\begin{equation}
    \label{eq_equilibrium}
    \mathrm{Div}P + f_b = 0,\ \mathrm{in} \ \mathcal{B}
\end{equation}
with displacement and traction boundary condition
\begin{equation}
    \label{eq_Dirichlet}
    \begin{split}
        u &= \bar{u} \ \mathrm{on} \ \Gamma_u, \\
        P\cdot N & = \bar{t} \ \ \mathrm{on} \ \Gamma_t.
    \end{split}
\end{equation}

\begin{figure*}[t]
    \centering
    \includegraphics[width=1\linewidth]{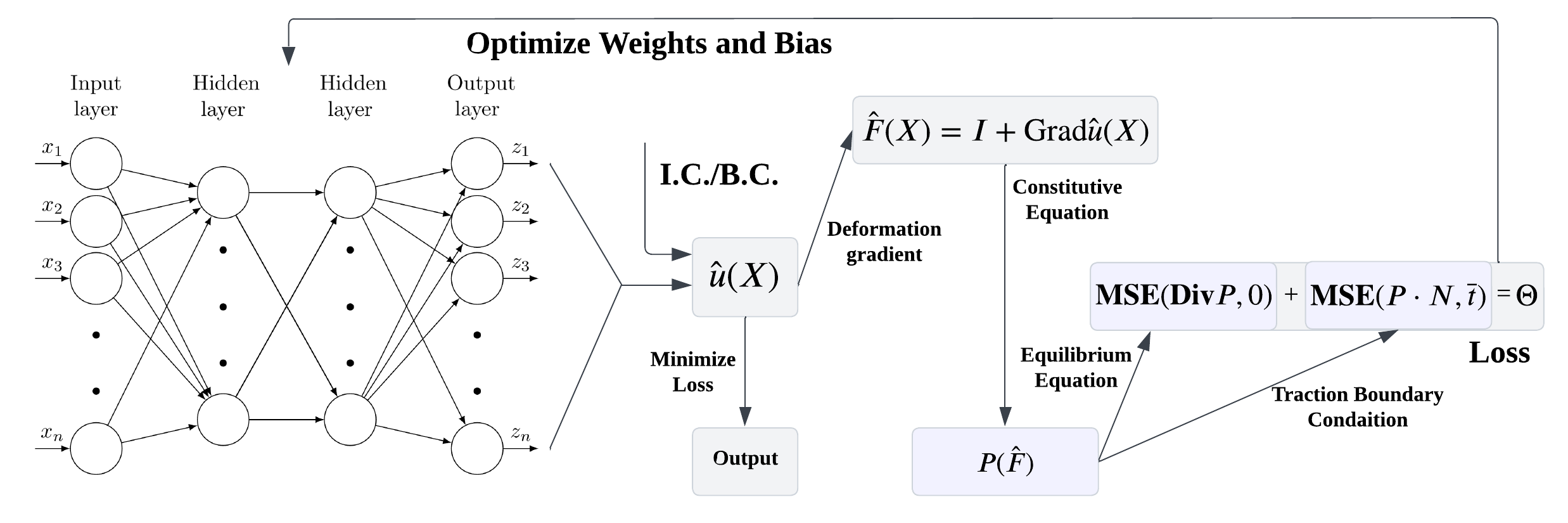}
    \caption{Physics-Informed Neural Network with Quasi-Static Condition.}
    \label{fig:PINN_net}
\end{figure*}

Figure~\ref{fig:PINN_net} illustrates the architecture of PINN. The total loss is computed using the equilibrium residual and the corresponding boundary conditions derived from the network architecture. Fuhg et al. \cite{fuhg2022mixed} discussed the PINN loss function with the following formulation:
\begin{equation}
    \label{eq_PINNtotalLoss}
    \begin{split}
        \Theta^* = \textbf{argmin}_\Theta & ( W_\mathcal{R} \mathrm{MSE}_\mathcal{R}(\{X^i_\mathcal{R}\}^{N_\mathcal{R}}_{i=1},\Theta) \\
        & + W_t \mathrm{MSE}_t (\{X^i_t\}^{N_t}_{i=1},\Theta) \\
        & + W_u \mathrm{MSE}_u (\{X^i_u\}^{N_u}_{i=1},\Theta)  )
    \end{split}
\end{equation}
where $W_\mathcal{R}, W_t$, and $W_u$ are the network weights, and $\{X^i_\mathcal{R}\}^{N_\mathcal{R}}_{i=1}$, $\{X^i_u\}^{N_u}_{i=1}$, and $\{X^i_u\}^{N_u}_{i=1}$ are the sets of domain, traction boundary, and displacement boundary particles, respectively.
The equilibrium loss can be calculated as
\begin{equation}
    \label{eq_PINNequilibriumLoss}
    \begin{split}
         \mathrm{MSE}_\mathcal{R}&(\{X^i_\mathcal{R}\}^{N_\mathcal{R}}_{i=1},\Theta) = \\
        &\sum\limits^{\mathrm{Dim}}_{j=1} \frac{1}{N_\mathcal{R}}\sum\limits^{N_\mathcal{R}}_{i=1}|\sum\limits_{k=1} \frac{\partial}{\partial X_k} P_{jk}(\hat{F}(X^i_\mathcal{R}, \Theta))|^2.
    \end{split}
\end{equation}
The traction and displacement boundary loss can be calculated as
\begin{equation}
    \label{eq_PINNboundaryLoss}
    \begin{split}
        \mathrm{MSE}_t (\{X^i_t\}^{N_t}_{i=1},\Theta) &= \frac{1}{N_t} \sum\limits^{N_t}_{i=1} |P(\hat{F}(X^i_t ,\Theta)\cdot N_i - \bar{t}(X^i_t))|^2, \\
        \mathrm{MSE}_u (\{X^i_u\}^{N_u}_{i=1},\Theta) & = \frac{1}{N_u} \sum\limits^{N_u}_{i=1} |\hat{u}(X^i_u , \Theta)-u(X^i_u)|^2.
    \end{split}
\end{equation}

The PINN loss function explicitly incorporates the traction boundary condition. This calculation performs a first-order derivative. It also requires computing second-order derivatives to satisfy the equilibrium condition. In the DEM method, as discussed in Section~\ref{sect:methodDEM}, the stationarity of the potential energy is enforced through an approximation that requires only first-order differentiation. This computational efficiency offers DEM a potential advantage in achieving faster model convergence.

\section{Conclusion}
\label{sect:conclusion}

In conclusion, we proposed an innovative neuro-symbolic framework for real-time solid deformation reconstruction, utilizing a NeRF and DEM. Our approach effectively utilizes neural networks and physical priors to solve the reconstruction task, highlighting the transformative potential of integrating neural learning with computational physics. By targeting real-time solid deformation reconstruction without requiring prior geometric or boundary information, our framework addresses key limitations in existing approaches, including reliance on neural networks for solid reconstruction and the need for low-cost computation.

Moreover, the implementation of this AI framework seamlessly integrates into existing manufacturing workflows, providing robust quality assurance without disrupting production lines.

As the field progresses, future research should focus on refining these methods to further enhance their generalizability, adaptability, and transferability to different materials and physics phenomena. Additionally, ongoing efforts shall aim to reduce the computational demands of these methods and explore their applicability across a wide range of physics simulation.

\section*{References}

\def\refname{}

\end{document}